\documentclass[10pt,twocolumn,letterpaper]{article}
\usepackage[margin=0.75in,top=0.9in]{geometry}
\usepackage{times}
\usepackage[T1]{fontenc}
\usepackage[utf8]{inputenc}
\usepackage{amsmath,amssymb,amsfonts}
\usepackage{graphicx}
\usepackage{booktabs}
\usepackage{pgfplots}
\pgfplotsset{compat=1.17}
\usepackage{multirow}
\usepackage{subcaption}
\usepackage{xcolor}
\usepackage{url}
\usepackage[pagebackref=false,breaklinks=true,colorlinks,citecolor=blue,linkcolor=red,urlcolor=blue]{hyperref}
\usepackage[capitalize]{cleveref}
\usepackage{microtype}
\usepackage{enumitem}
\usepackage{multicol}
\setlist{nosep,leftmargin=*}
\newcommand{\ours}{\textsc{Ours}}
\newcommand{\chcat}{channel-wise concatenation}
\newcommand{\seqcat}{sequence concatenation}
\newcommand{\Chcat}{Channel-wise concatenation}

\newcommand{\ie}{\emph{i.e.}}
\graphicspath{{figures/}}

\title{Latent Fusion: Channel Stacking for Structure-Preserving Image Editing}
\author{Mohit Dulani\\ \texttt{mohitdulani28@gmail.com}}
\date{}

\begin{document}
\maketitle

\begin{abstract}
\small
Instruction-based image editing models must condition on the source image, and transformer-based editors do so in one of two ways: sequence concatenation, which appends the source-image tokens to the target sequence, as in FLUX.1 Kontext, or channel-wise concatenation, which stacks the source latent onto the noisy latent and widens the input projection, as in InstructPix2Pix and FLUX.1 Fill. The FLUX.1 Kontext report found channel-wise concatenation to perform worse in initial experiments, conducted at 1K resolution with a 12B model and a large editing corpus. In this work we explore channel-wise concatenation on a Flux backbone as the lower-cost option for a regime that is common in production: structure-preserving edits that change appearance (sky, exposure, foliage, cleanup) while leaving the layout and identity of the source untouched, about 1{,}400 training pairs (a small subset), a 4B non-distilled backbone, and 2K frames. Channel-wise concatenation leaves the sequence at its original length, where sequence concatenation doubles it and roughly quadruples attention cost, and it needs only standard single-scale classifier-free guidance, two forward passes per step, where the InstructPix2Pix two-scale scheme needs three. Starting from FLUX.2 Klein 4B base, we widen only the input projection so that every image token carries its own source content, and we train both conditioning schemes with the same trainer, data, and hyperparameters. We observe that channel-wise concatenation keeps the output pixel-aligned with the source and changes its appearance without moving its structure; in our runs sequence concatenation, which must learn the correspondence between 10{,}880 target tokens and 10{,}880 source tokens from fewer than 1{,}500 pairs, displaced structure on this data, an observation we report without claiming a general ranking. On 29 held-out images the editor reaches 20.3 dB PSNR and 0.198 LPIPS against the human edit, where the unedited source scores 15.0 dB, and counterfactual conditioning tests show that what it learned is largely a global grade of the source rather than a selective edit, which we report as the open problem of this recipe. We also compare against an SD3.5-Medium editor trained with the same surgery and report the full training recipe, including the failure modes encountered along the way. Code is available at \url{https://github.com/complete-dope/LatentFusion}.

\end{abstract}

\section{Introduction}
\label{sec:intro}

Instruction-based image editing~\cite{brooks2023instructpix2pix} has become a standard application of generative models: given a source image and an instruction, a conditional diffusion model produces the edited image with no per-image optimisation. Most benchmarks in this area reward content-changing edits, where the instruction asks for new objects or a rearranged scene and some deviation from the source is acceptable. In many production settings, however, the edit is conservative. A photograph is submitted, and the expected output is the same photograph with a replaced sky, a corrected exposure, healthier foliage, or a distraction removed. The buildings, streets, furniture, and people must remain pixel-aligned with the source, because the result is inspected side by side with the original and any drift is a defect. The two settings place different demands on how the source image is fed to the network.

Transformer-based editors condition on the source image in one of two ways. FLUX.1 Kontext~\cite{batifol2025kontext} and most recent editors use \seqcat{}: the source-image tokens are appended to the target sequence and distinguished by a positional offset. This design lets the network read the source from any position, which suits content generation, but the correspondence between source and target tokens must be learned through attention, and the sequence length doubles. InstructPix2Pix~\cite{brooks2023instructpix2pix} and FLUX.1 Fill~\cite{bfl2024fluxfill} instead use \chcat{}: the source latent is stacked onto the noisy latent along the channel dimension and the first layer of the network is widened. Each token then carries its own source content, the sequence length is unchanged, and the source-target correspondence is fixed by construction. The Kontext report notes that \chcat{} performed worse in their initial experiments, conducted at 1K resolution with a 12B model and a large editing corpus.

In this work we explore \chcat{} on a Flux backbone as the lower-cost conditioning for the structure-preserving regime, and we call the resulting channel-stacked editor \textsc{Latent Fusion}. The motivation is cost, not a rejection of \seqcat{}: \chcat{} keeps the sequence at $L$ tokens where \seqcat{} needs $2L$, so attention cost stays that of the base model, and at inference it needs only standard classifier-free guidance, two forward passes per step, where the InstructPix2Pix two-scale guidance needs three (\cref{sec:surgery}). Our setting differs from the Kontext comparison on every axis: the edit must leave structure and identity intact, the training set consists of 1{,}398 pairs from a single professional workflow, the backbone is the 4B non-distilled FLUX.2 Klein base model~\cite{bfl2026flux2klein}, and the frames are $2048\times1360$, \ie{} $10{,}880$ image tokens. We implement both conditioning schemes in the same trainer and train both on identical data with identical hyperparameters. We observe that \chcat{} preserves object placement and scene structure and changes appearance without moving anything. In our run of \seqcat{}, with the same backbone and data, the source content itself moved: objects shifted and structure was regenerated rather than preserved, so the output was no longer registered to the source (\cref{sec:results}). We report this as what we saw on this data with a single run of each scheme, not as evidence that \seqcat{} is the weaker design on Flux.

Our contributions are as follows:
\begin{itemize}
  \item A recipe for adapting the FLUX.2 Klein 4B base model into a paired editor with \chcat{} conditioning: only the input projection is widened, its new half is zero-initialised and the layer is trained with its own learning rate, and training mixes native-resolution crops with downscaled whole frames under conditioning dropout. All hyperparameters are reported (\cref{sec:method}).
  \item A matched comparison of \chcat{} against \seqcat{} on the same backbone, data, and hyperparameters in the structure-preserving regime, by inspection of held-out outputs, with full-frame metrics against human reference edits for the channel-wise editor, together with the compute and guidance cost of the two schemes (\cref{sec:surgery,sec:results}).
  \item A comparison against an SD3.5-Medium editor trained with the same surgery at the same resolution, and an analysis of what the widened input layer learned, counterfactual conditioning tests that expose copying rather than editing, and a cross-architecture control on SD3-Medium (\cref{sec:ablations}).
  \item A record of the training failures we encountered and how each was diagnosed (\cref{sec:appendix}).
\end{itemize}
Code for the trainer, both conditioning paths, the sampler and the evaluation scripts is released in the \textsc{Latent Fusion} repository.

\Chcat{} on a Flux backbone is not new; FLUX.1 Fill uses it for inpainting. Our claim is narrower and concerns the regime: when the edit must preserve the source and paired data are scarce, fixing the source-target correspondence by construction is enough to obtain a working editor at the token count and attention cost of the base model, with two-pass guidance rather than the three-pass InstructPix2Pix scheme, and without the doubled sequence that \seqcat{} requires.

\section{Related Work}
\label{sec:related}

\paragraph{Instruction-based and paired image editing.}
Paired-edit models differ mainly in how the source image enters the network.
InstructPix2Pix~\cite{brooks2023instructpix2pix} concatenates the clean source latent to the noisy latent along the channel axis and widens the first convolution of a Stable Diffusion 1.5 U-Net; MagicBrush~\cite{zhang2023magicbrush} fine-tunes that same InstructPix2Pix model on human-annotated triplets.
Emu Edit~\cite{sheynin2024emuedit} uses the same channel-concatenation surgery on Meta's Emu backbone, adding input channels with zero-initialised weights, and supervises it with task embeddings.
UltraEdit~\cite{zhao2024ultraedit} scales the data to roughly 4M triplets and trains a channel-concatenation editor on SD3.
FLUX.1 Fill~\cite{bfl2024fluxfill} applies the same construction to the 12B FLUX.1 rectified-flow transformer~\cite{bfl2024flux1}, concatenating a masked image and a binary mask for inpainting.
The alternative is to append the conditioning image as extra tokens.
FLUX.1 Kontext~\cite{batifol2025kontext} appends the source-image tokens to the target sequence inside FLUX.1, and tells the two apart by offsetting their 3D RoPE positions rather than by any extra input channel; its report states that channel concatenation was the weaker of the two in their early experiments.
ICEdit~\cite{zhang2025icedit} builds an in-context editor on FLUX.1 Fill with LoRA~\cite{hu2022lora}, and OmniGen~\cite{xiao2024omnigen} embeds image patches as visual tokens in a single Phi-3-initialised transformer with the SDXL VAE. Step1X-Edit~\cite{liu2025step1xedit} pairs an MLLM with a DiT decoder, Qwen-Image-Edit~\cite{wu2025qwenimage} conditions its own MMDiT on both semantic and reconstructive image encodings, and BAGEL~\cite{deng2025bagel} uses a Qwen2.5-initialised mixture-of-transformer-experts over FLUX VAE latents.
DiT-VTON~\cite{li2025ditvton} compares token concatenation, channel concatenation and ControlNet conditioning, but only for virtual try-on and on an undisclosed DiT.
The two conditioning schemes have not, to our knowledge, been compared for paired editing on a common Flux backbone under matched data and hyperparameters.

\paragraph{Why the regime matters.}
The Kontext comparison was made with the resources of a large lab: a 12B model and a corpus large enough to learn any correspondence attention needs. Under sequence concatenation the correspondence between a target token and its source token is not given to the network; it has to be learned from data, through attention, over a sequence twice as long. Under channel concatenation it is fixed by construction. Which of the two is better should therefore depend on how much data there is to learn the correspondence from, and on whether the task even wants a free correspondence. Content-changing edits do; structure-preserving edits do not. The two also differ in cost: sequence concatenation doubles the token count, and the InstructPix2Pix guidance scheme usually paired with channel concatenation costs three passes per step where a single text guidance scale costs two. None of the works above vary these factors while holding the backbone fixed. We do.

\paragraph{Backbone.}
Rectified-flow transformers~\cite{esser2024sd3,liu2023rectifiedflow,lipman2023flowmatching} with rotary position embeddings~\cite{su2021roformer} are the current standard for image generation; DiT~\cite{peebles2023dit} fixed the patch-token layout they use. FLUX.2 Klein 4B~\cite{bfl2026flux2klein} is a small member of this family, released in both a distilled and a non-distilled (base) form. We use the base form; our distilled-checkpoint runs are reported as indicative only.

\paragraph{Positioning.}
We train InstructPix2Pix-style channel-concatenation conditioning on a non-distilled FLUX.2 Klein 4B base model for structure-preserving paired editing at $2048\times1360$, on about 1{,}400 pairs, and compare it against sequence-concatenation conditioning on the same backbone, data and hyperparameters, and against an SD3.5-Medium editor with the same surgery. We present channel concatenation as the cheaper of the two schemes for this regime, not as a replacement for sequence concatenation in general.

\section{Method}
\label{sec:method}

\subsection{Problem}
Given a source photograph $I_s$ and a tag prompt $p$, produce the edited photograph $I_e$ at $2048\times1360$. Training pairs $(I_s, I_e, p)$ come from a professional editing workflow; $I_e$ is a human-finished edit of $I_s$, so the two are pixel-aligned, and the edit changes appearance (sky, exposure, foliage, cleanup) but not the layout or identity of what is in the frame. This is the InstructPix2Pix setting \cite{brooks2023instructpix2pix} with paired supervision, restricted to structure-preserving edits.

\subsection{Backbone}
We use FLUX.2 Klein 4B \emph{base}, the non-distilled checkpoint \cite{bfl2026flux2klein}. It is a rectified-flow transformer with 5 double-stream and 20 single-stream blocks, hidden size 3072, and 24 attention heads. Positions are encoded with rotary embeddings over four axes \cite{su2021roformer}. The text encoder is a frozen Qwen3-4B. The VAE shrinks the image by $8\times$ in each direction and produces 32 channels. Each $2\times2$ block of latent pixels is then folded into one token, so a token covers a $16\times16$ pixel block and has 128 channels. A $2048\times1360$ frame is $128\times85 = 10{,}880$ image tokens.

We first tried the distilled Klein 4B checkpoint, at 512 pixels with LoRA and without channel concatenation; its edits did not meet our fidelity bar. \Cref{sec:ablations} reports those runs as indicative only; no matched comparison exists. All results below use the base checkpoint.

\subsection{Two ways to condition on the source}
\label{sec:surgery}
Let $x_0 \in \mathbb{R}^{L\times128}$ be the token sequence of the edited image $I_e$ and $c \in \mathbb{R}^{L\times128}$ that of the source image $I_s$, both normalised with the VAE's stored per-channel mean and variance, with $L=10{,}880$.

\paragraph{Channel concatenation (ours).}
The only change to the network is to its first layer, the input projection \texttt{x\_embedder}. It is a linear layer without bias, $W\in\mathbb{R}^{3072\times128}$. We replace it with $W' \in \mathbb{R}^{3072\times256}$,
\begin{equation}
W' = [\,W \;\; \mathbf{0}\,],
\end{equation}
that is, the pretrained weights are kept and the new half is set to zero. At the start of training the model therefore ignores $c$ and behaves exactly like the pretrained one. Each input token is the noisy token stacked with its source token, $[\,x_t \,\|\, c\,]$. The number of tokens does not change, the output layer does not change and still predicts 128 channels, and $c$ is never noised. Token $i$ of the input holds the source pixels of exactly the $16\times16$ block it will predict; the correspondence between source and target is fixed by construction and costs no attention. InstructPix2Pix does the same thing on a U-Net and FLUX.1 Fill does it on a Flux transformer for inpainting; we apply it to paired editing on the base model.

\paragraph{Sequence concatenation (reference scheme).}
The alternative, used by FLUX.1 Kontext \cite{batifol2025kontext} and most recent transformer editors, leaves the input projection unchanged and appends the tokens of $c$ after those of $x_t$, giving a sequence of $2L = 21{,}760$ tokens. The source tokens receive the same spatial positions as their target counterparts and an offset on a non-spatial RoPE axis, so the two images can be told apart. The network reads the source through attention: every target token can attend to every source token, and it has to learn from data that the useful one is the token at its own position. Only the $L$ target positions are supervised; the source positions produce outputs that are discarded. The Kontext report states that channel-wise concatenation performed worse than sequence concatenation in their initial experiments, at 1K resolution with a 12B model and a large edit corpus. We implement both schemes in the same trainer, behind a single flag, and everything else in this section applies to both. 

\paragraph{Cost.}
With $L$ image tokens, channel-wise concatenation leaves the sequence at $L$ and widens only the input projection, so attention cost, which grows as $L^2$, activation memory and step time stay those of the base model. Sequence concatenation doubles the sequence to $2L$, or $(k+1)L$ for $k$ reference images: attention FLOPs per layer rise about four times and the linear layers twice, so a step costs roughly two to four times the compute and activation memory. Here $L=10{,}880$. Guidance adds a second axis of cost, discussed below: channel-wise concatenation with a single text guidance scale needs two forward passes per sampling step, while the InstructPix2Pix two-scale scheme needs three.

\subsection{Training objective}
We use the standard rectified-flow objective \cite{esser2024sd3,liu2023rectifiedflow,lipman2023flowmatching}: the network predicts the straight-line velocity from noise to the clean image. With noise $\epsilon\sim\mathcal{N}(0,I)$ and noise level $\sigma_t\in(0,1]$,
\begin{align}
x_t &= (1-\sigma_t)\,x_0 + \sigma_t\,\epsilon, \\
\mathcal{L} &= \mathbb{E}_{t,\epsilon}\,\bigl\|\, v_\theta(x_t,\, c,\, t,\, E_p) - (\epsilon - x_0)\,\bigr\|_2^2 ,
\end{align}
where $v_\theta(x_t,c,\cdot)$ is $v_\theta([x_t\,\|\,c],\cdot)$ under channel concatenation and the first $L$ outputs of $v_\theta([x_t;\,c],\cdot)$ under sequence concatenation. $E_p$ is the prompt embedding. Noise levels are drawn uniformly from the checkpoint's schedule and every level is weighted equally. Following InstructPix2Pix, we sometimes hide the conditioning during training, with $q=0.05$: with probability $q$ the text is replaced by the empty prompt, with probability $q$ the source image is zeroed, and with probability $q$ both are. This makes it possible to use two separate guidance scales at inference,
\begin{equation}
\label{eq:cfg}
\tilde v = v(\varnothing,\varnothing) + g_i\bigl[v(c,\varnothing)-v(\varnothing,\varnothing)\bigr] + g_t\bigl[v(c,p)-v(c,\varnothing)\bigr],
\end{equation}
where $g_i$ controls how strongly the output follows the source image and $g_t$ how strongly it follows the text, at the cost of three forward passes per step. Because the source enters through the input channels and is never noised, the image term is not needed to keep the output on the source; dropping it and keeping a single text scale $g_t>1$ reduces \cref{eq:cfg} to standard classifier-free guidance, $\tilde v = v(c,\varnothing) + g_t[v(c,p)-v(c,\varnothing)]$, at two passes per step. Our SD3.5-M baseline uses the three-pass form of \cref{eq:cfg}. The Flux editors reported here use neither: Klein 4B has no guidance embedding, and our sampler runs one conditional pass per step; no unconditional or image-only pass is computed, so every output of the main run is unguided (the follow-up's held-out scoring in \cref{sec:ablations} is the exception). 

\subsection{Data pipeline and resolution}
\label{sec:datapipe}
Each training pair is stored at $4096\times2728$. At every step, with probability 0.5 we cut a random $2048\times1360$ window out of the frame at full pixel scale; otherwise we shrink the whole frame by exactly $2\times$ to $2048\times1364$ and trim 4 rows to reach $2048\times1360$. The source and the edit get the same cut, so their tokens stay aligned pixel for pixel, and window offsets are multiples of the 16-pixel token size. The network therefore sees fine detail at native scale in half of its steps and the whole layout in the other half, always as $10{,}880$ tokens. We do not flip images horizontally.

\subsection{Optimisation}
We fine-tune every weight of the transformer; the VAE and text encoder stay frozen. The optimiser is AdamW with $\beta=(0.9,0.999)$ and weight decay $10^{-4}$. The learning rate is $10^{-5}$ for the transformer body and $10^{-3}$ for the widened input layer, because its new half starts at zero and has to grow. The rate warms up over 400 steps and then follows a cosine decay over 20{,}000 steps. Gradients are clipped at norm 1.0. Each of 3 H100-80GB GPUs processes 8 crops per micro-batch and accumulates 3 micro-batches per update, for 72 crops per optimiser step. Compute runs in bf16 with fp32 master weights, and the model is sharded across GPUs with FSDP. 
The \seqcat{} run uses the same settings with the conditioning flag flipped; its input projection is not widened, so it has no separate learning-rate group.
The \chcat{} model was trained for 20{,}000 steps on 3 H100-80GB GPUs. At 41 s per optimiser step this is about 230 hours of wall-clock time, or roughly 680 GPU-hours. No training log survives for the \seqcat{} run, so its step count and wall time are not reported (\cref{sec:limitations}). 

\subsection{Inference}
\label{sec:infer}
At inference the $2048\times1360$ source is encoded once into $c$ and the output is denoised in a single pass with $N=28$ Euler steps, the noise level going from $1$ down to $1/N$, and the checkpoint's schedule shift at its value for $L=10{,}880$. Under channel concatenation the sequence is $L$ tokens per step; under sequence concatenation it is $2L$, so each step costs about four times the attention compute and twice the rest (\cref{sec:surgery}). 

\section{Experiments}
\label{sec:experiments}

\subsection{Data}
\label{sec:data}
The dataset has 1{,}427 training pairs and a 24-pair test split from one professional editing workflow. We hold out 2\% of the training pairs (29, fixed seed) for evaluation and train on the remaining 1{,}398; the test split is not trained on and supplies the pairs for the counterfactual tests in \cref{sec:inputlayer}. Each pair is a raw capture and its human-finished edit (sky, ground and foliage, exposure and white balance, structural cleanup). All frames are landscape. 180 training pairs were edited entirely by hand, without any AI tool; the remaining edits were AI-assisted and human-finished. Source frames are $4096\times2728$; the model trains on $2048\times1360$ windows and $2\times$ downscaled frames (\cref{sec:datapipe}). 
Prompts are short tag lists naming the target attributes rather than free-form instructions, of the form \texttt{clear sky, balanced exposure, green foliage}; the four tags in the data cover sky, exposure, foliage and structure, and the tag vocabulary itself is not released with the data.

\subsection{Held-out evaluation set}
\label{sec:evalset}
The held-out set is the 2\% split above: 29 pairs drawn with a fixed seed and excluded from training. Each pair is evaluated at $2048\times1360$ from a $2\times$ downscale of the source, with one output per image, 28 sampling steps, and the seed set to the image index.\footnote{For the main run's checkpoints, the guidance value recorded in the run configuration is inert: the base checkpoint has no guidance embedding and our sampler applies none, so every reported output is unguided.} Each has a human ground-truth edit. Metrics on this set for every saved checkpoint are in \cref{tab:ckpt,fig:heldout}. 

\subsection{Metrics}
\label{sec:metrics}
We report PSNR, SSIM, and LPIPS \cite{zhang2018lpips} between each output and the human edit over the whole frame.

\subsection{Baselines}
\begin{itemize}
  \item \textbf{Source (no edit).} The unedited source, as a floor for every metric.
  \item \textbf{SD3.5-M IP2P.} An SD3.5-Medium editor with the same channel-concatenation surgery, trained at $2048\times1360$ on the same data.
  \item \textbf{Klein 4B base, \seqcat{}.} The same trainer, data and hyperparameters as ours, with the source appended as tokens (\cref{sec:surgery}).
  \item \textbf{Klein 4B base, pretrained \seqcat{} path.} The pretrained model, no fine-tuning, driven by a natural-language instruction through its own image-conditioning path, with an image-guidance sweep on one frame (\cref{sec:results}).
  \item \textbf{\ours.} FLUX.2 Klein 4B base, \chcat{}.
\end{itemize}
Held-out metrics are reported for \ours{} and the source; the other three are compared qualitatively in \cref{sec:results}. The distilled-checkpoint variant and the recipe ablations are reported in \cref{sec:ablations}.

\subsection{Implementation}
Our model's training settings are in \cref{sec:method}. The SD3.5-Medium baseline is a full fine-tune with the same channel-stacking change, on 16-channel latents (16 noisy + 16 source). It uses logit-normal noise-level sampling, learning rate $10^{-5}$, batch 1 with 4 accumulation steps, 500 warmup steps, and 20{,}000 steps in bf16; it took 38.4 hours on 3 RTX PRO 6000 GPUs. It is sampled with \cref{eq:cfg} at $g_t=2.4$, $g_i=1.4$; these values were fixed by hand and not swept on this checkpoint. A sweep on a separate 4K SD3.5-M checkpoint preferred $g_i=1.2$ by about 0.2 dB, so the baseline may be slightly under-tuned. 

\section{Results}
\label{sec:results}

\begin{figure*}[t]
\centering
\includegraphics[width=\textwidth]{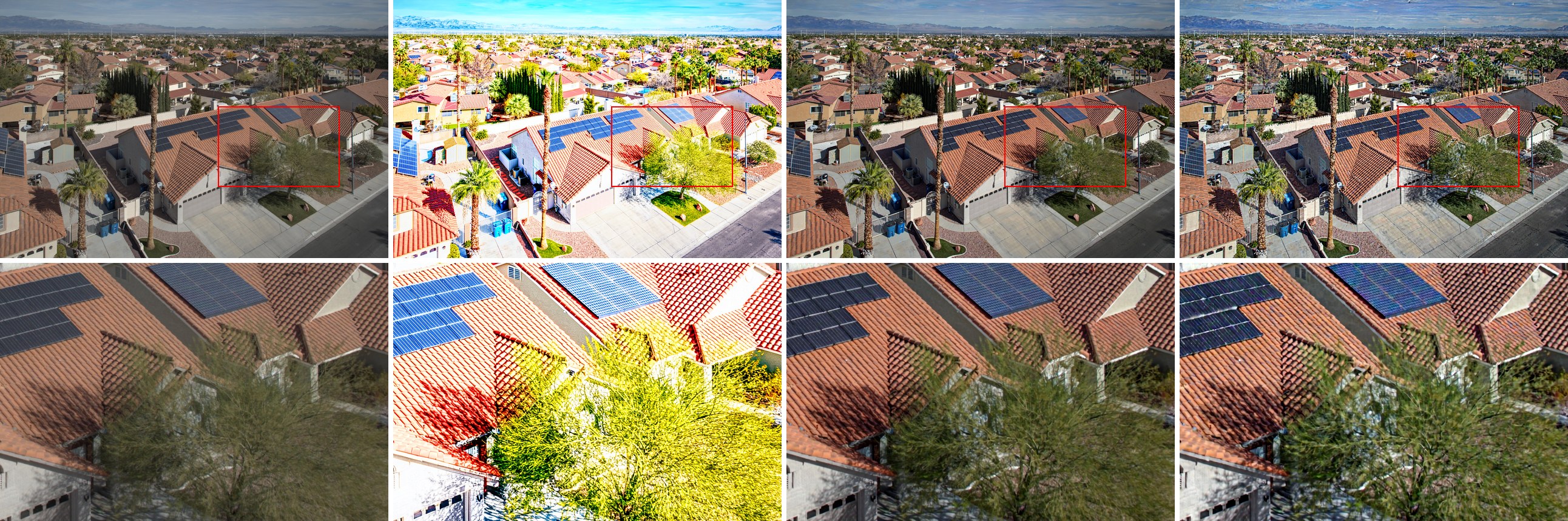}
\caption{The pretrained Klein 4B base model through its own \seqcat{} editing path, without fine-tuning, on one $2048\times1360$ frame with the instruction to keep the scene unchanged and brighten, white-balance and add contrast. Columns: source, then image-guidance scale 1, 5 and 10 at text guidance 4, 50 steps, seed 0. Bottom row: the boxed region at 100\%. At scale 1 the source is largely ignored and the colours blow out; at 5 and 10 the output is pixel-registered to the source and the edit is a mild brightening.}
\label{fig:seqcat_guidance}
\end{figure*}

\begin{table*}[t]
\centering\small
\setlength{\tabcolsep}{5pt}
\caption{Held-out set (29 images) at $2048\times1360$, \ours{} at step 19{,}500. Full-frame metrics against the human edit, mean $\pm$ std over images. The source row is the unedited input; its LPIPS was not computed.}
\label{tab:main}
\begin{tabular}{lccc}
\toprule
Method & PSNR$\uparrow$ & SSIM$\uparrow$ & LPIPS$\downarrow$ \\
\midrule
Source (no edit)                  & $14.98 \pm 1.97$ & $0.553 \pm 0.193$ & -- \\ 
\ours{} (Klein 4B base, \chcat{}) & $\mathbf{20.29 \pm 3.02}$ & $\mathbf{0.720 \pm 0.143}$ & $\mathbf{0.198 \pm 0.050}$ \\ 
\bottomrule
\end{tabular}
\end{table*}

\begin{figure*}[t]
\centering
\includegraphics[width=\textwidth]{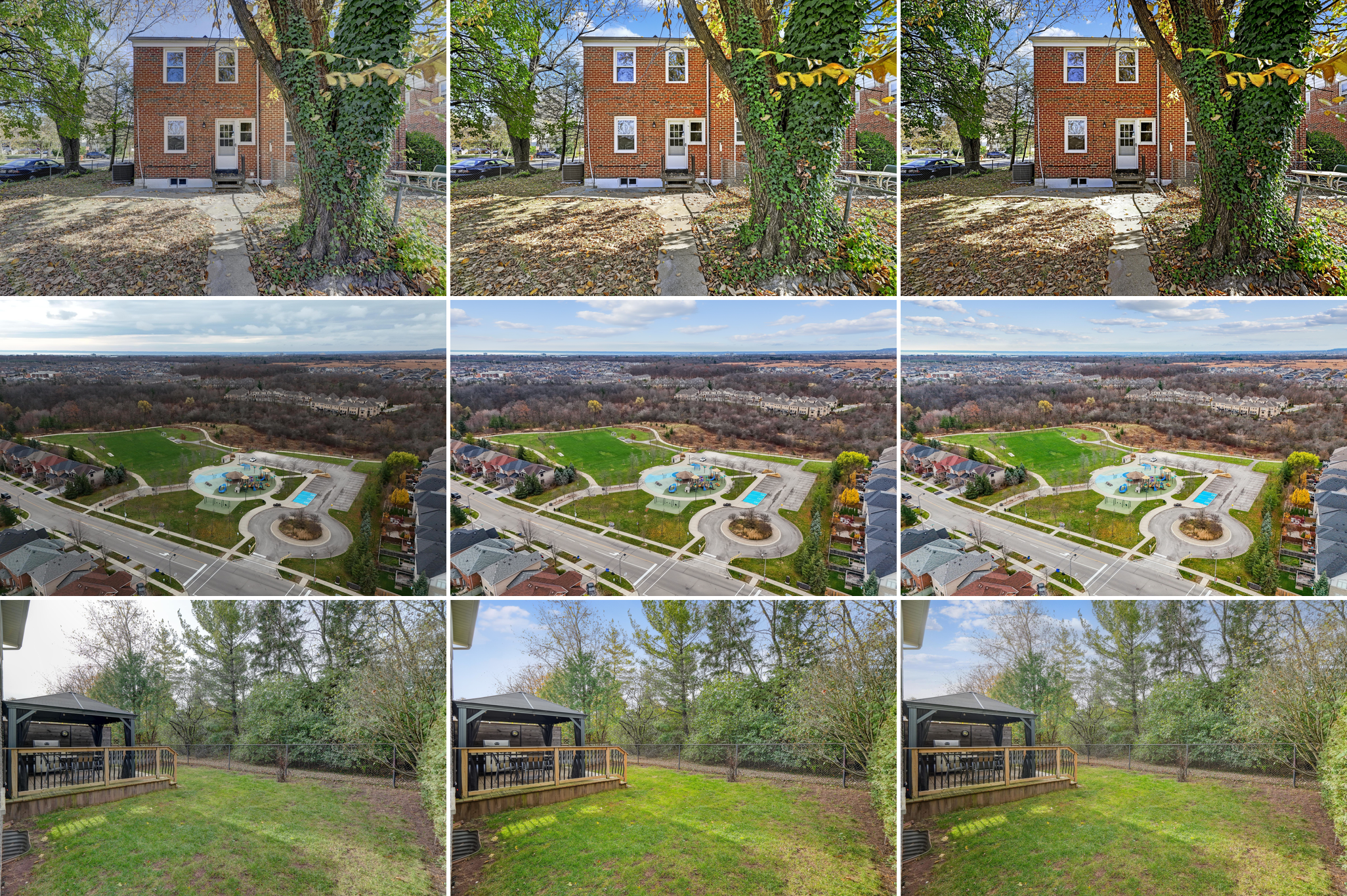}
\caption{Held-out results at step 19{,}500. Columns: source, \ours{}, human edit. Rows are the held-out images at the 25th, 50th and 75th percentile of PSNR, not hand-picked. The edits change sky, exposure and foliage; buildings, roads, fences and furniture stay in place. Frames are downscaled from $2048\times1360$. }
\label{fig:qual} 
\end{figure*}

\paragraph{Channel against sequence concatenation.}
We trained both conditioning schemes on the same data with the same trainer and hyperparameters and compared their outputs on held-out images by inspection at 100\% zoom. With sequence concatenation the source content itself moved: objects shifted and structure was rebuilt rather than kept, so the output no longer registered against the source pixel for pixel. With channel-wise concatenation the appearance changed and everything held its position, which is what the task needs, since the human reference edit is pixel-aligned with the source. The sequence-concatenation model was not scored on the held-out set, so \cref{tab:main} reports metrics for \ours{} alone; the comparison between the two schemes rests on this visual inspection of one run each. We do not read it as a rejection of sequence concatenation on Flux, which remains the established design for content-changing edits; the point of this paper is that channel-wise concatenation, at the token count and guidance cost of the base model, is sufficient for the structure-preserving case. On the 29 held-out images \ours{} reaches 20.3 dB PSNR, 0.720 SSIM and 0.198 LPIPS against the human edit, where the unedited source scores 15.0 dB and 0.553 SSIM; PSNR improves over the source on 28 of the 29 images. \Cref{fig:qual} shows the three images at the quartiles of PSNR. The gain over the source is largely a global grade: sky, exposure and foliage move toward the human edit across the frame, and the structure of the scene is untouched. The counterfactual tests in \cref{sec:inputlayer} show that this is also the limit of what the model learned; it reproduces the source it is handed and grades it, rather than reading the source and changing it selectively. 

\begin{table}[t]
\centering\small
\caption{\ours{} on the 29 held-out images at selected checkpoints, mean over images. $\rho$ is the conditioning-weight ratio of \cref{sec:ablations}. All 36 checkpoints are plotted in \cref{fig:heldout}.}
\label{tab:ckpt}
\begin{tabular}{rccc}
\toprule
Step & PSNR$\uparrow$ & LPIPS$\downarrow$ & $\rho$ \\
\midrule
2{,}000  & 18.43 & 0.227 & 1.06 \\
5{,}000  & 19.97 & 0.206 & 1.20 \\
10{,}000 & 20.30 & 0.202 & 1.29 \\
15{,}000 & 20.25 & 0.199 & 1.31 \\
19{,}500 & 20.29 & 0.198 & 1.31 \\
\bottomrule
\end{tabular}
\end{table} 

\begin{figure*}[t]
\centering
\includegraphics[width=\textwidth]{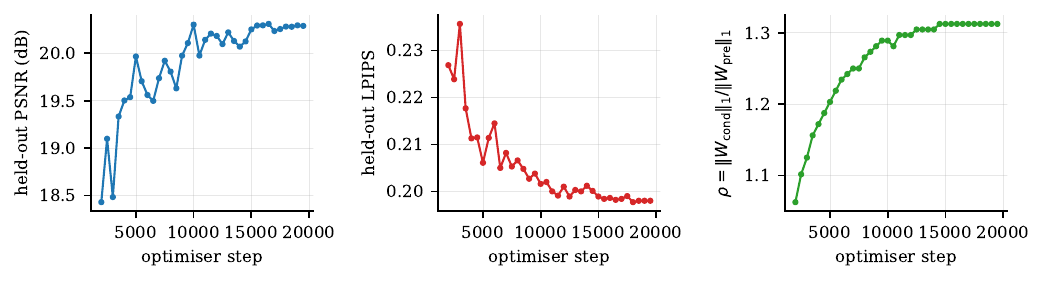}
\caption{\ours{} on the 29 held-out images at every checkpoint (every 500 steps). Left: PSNR against the human edit. Middle: LPIPS. Right: the weight ratio $\rho$ of the widened input projection. Both metrics plateau by step 10{,}000; $\rho$ saturates at 1.31 by step 14{,}500.}
\label{fig:heldout}
\end{figure*}

\paragraph{Held-out metrics over training.}
\Cref{tab:ckpt,fig:heldout} track the held-out set across the 36 checkpoints. PSNR rises from 18.4 dB at step 2{,}000 to 20.3 dB at step 10{,}000 and LPIPS falls from 0.227 to 0.202; both are flat over the second half of training, as is the training loss (0.568 averaged over steps 9{,}500--10{,}500, 0.564 over the last 1{,}000 steps), and $\rho$ stops growing at 1.31. The model does not overfit the held-out set, and the second half of training buys about 0.004 LPIPS.

\paragraph{The pretrained \seqcat{} path.}
Klein 4B base ships with a \seqcat{} editing path of its own, so we also ran the pretrained model without fine-tuning on one $2048\times1360$ frame, with a natural-language instruction to keep the scene unchanged and brighten it, sweeping the image-guidance scale over 1, 5 and 10 at text guidance 4 and 50 steps (\cref{fig:seqcat_guidance}). This frame is not in the held-out set and has no human reference, so we measure only how far each output departs from its source. At image guidance 1 the output is essentially a new render in the same layout: 8.3 dB PSNR and 0.43 SSIM against the source, with saturated colour across the frame. At 5 and 10 the output stays registered to the source, with 24.4 and 19.9 dB PSNR and 0.88 and 0.73 SSIM against it, and phase correlation on nine $256\times256$ patches measures no shift at either scale; the edit is a mild global brightening. The pretrained path therefore preserves structure once image guidance is high enough, at the cost of three forward passes per step over $2L$ tokens, and it does not perform the domain-specific edit that the paired data encode. The structural drift we observed under \seqcat{} (previous paragraph) was in the fine-tuned run, not in the pretrained path. 

\paragraph{Against the SD3.5-Medium editor.}
The SD3.5-Medium editor with the same surgery (16-channel latent, 2.5B MMDiT) produced edits of comparable quality to \ours{} on this data; we did not find a consistent gap between the two on inspection, and we did not score it on the held-out set. We chose Klein over SD3.5-Medium for its position encoding rather than for quality: SD3.5-Medium adds a fixed-size two-dimensional sinusoidal position table to its patches~\cite{esser2024sd3}, which caps the token grid it can address, whereas Klein encodes position with rotary embeddings that extend to any grid, so the same weights serve the $2048\times1360$ frames here and larger ones later. 

\section{Ablations}
\label{sec:ablations}

Each row of \cref{tab:overfit} is a single training run; we did not repeat runs with different seeds. Run identifiers are recorded in the source comments.

\paragraph{Distilled vs base.}
The existing distilled runs are 512 px, LoRA rank 4, without channel concatenation, so they are not a matched pair with the base run; \cref{tab:overfit} reports them as indicative only. The fidelity failure we saw with the distilled checkpoint was on one crop, before any of the recipe was in place.

\begin{figure*}[t]
\centering
\includegraphics[width=\textwidth]{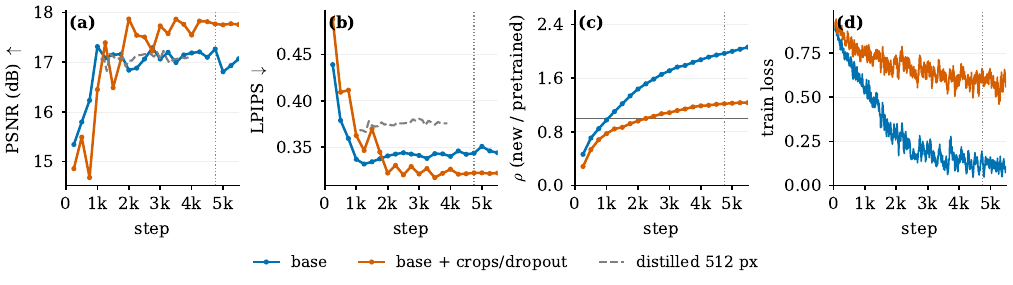}
\caption{Subset runs (\cref{tab:overfit}), validation metrics every 250 steps and training loss, cut to the first 5{,}500 steps, the length of the crops run. (a) PSNR of every base run peaks within the first 2K steps from the pretrained weights and does not improve after. (b) LPIPS. (c) The conditioning-weight ratio $\rho$ starts at zero and passes 1 (thin line) between 1{,}250 and 2{,}500 steps, so the widened projection is used; runs without crops keep pushing $\rho$ up while validation quality is flat. The base curve is run A continued by run B from A's checkpoint at step 4{,}750 (dotted line). The distilled 512\,px run is dashed grey, indicative only, and has no $\rho$ because its input layer is not widened. (d) Training loss, 50-step running mean: crops and dropout keep it high while whole-frame training memorises the 16 pairs; the crops run also uses warmup and cosine decay, which would lower the loss rather than raise it. Run B beyond step 5{,}500 is in \cref{tab:overfit}.}
\label{fig:subset}
\end{figure*}

\begin{figure}[t]
\centering
\includegraphics[width=\linewidth]{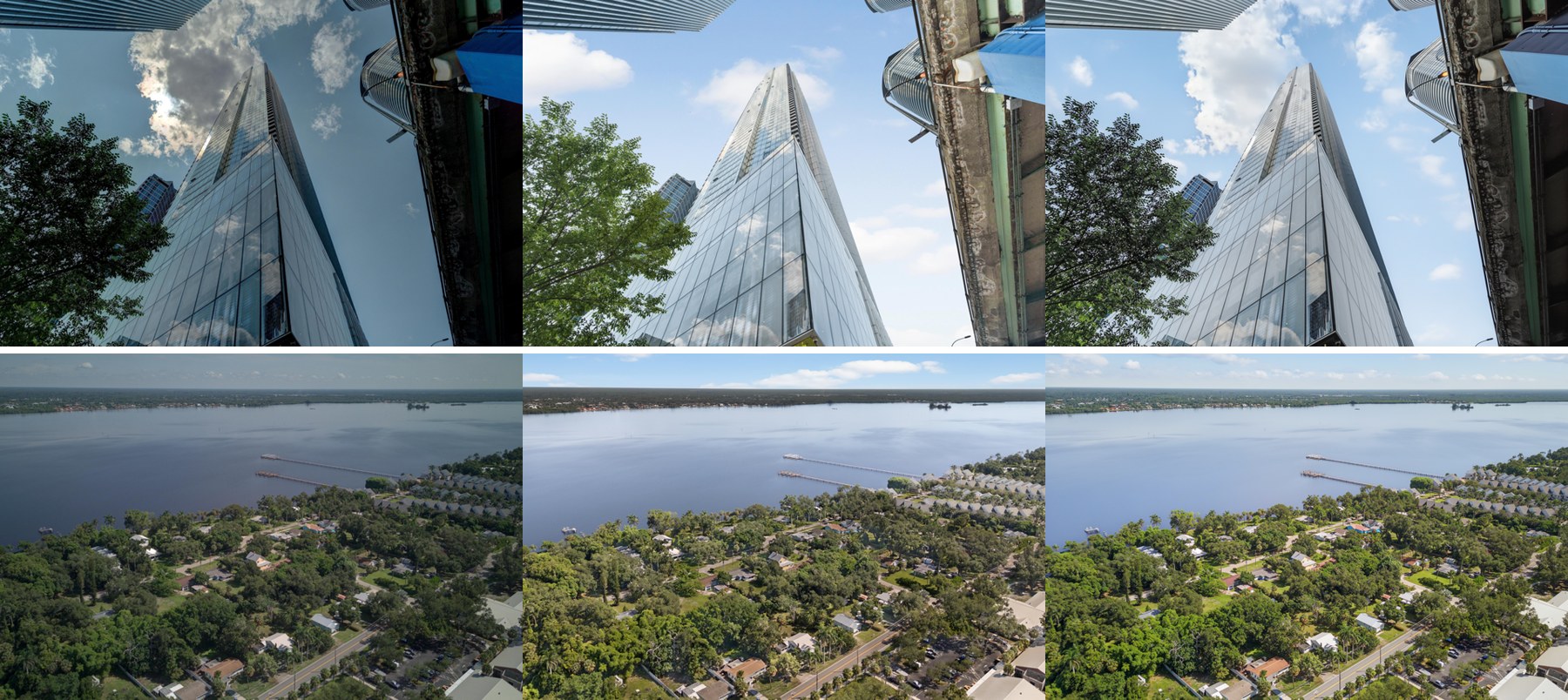}
\caption{Two of the four validation pairs of the subset run with crops, at its best-PSNR step (2000). Each row: source, human edit, our output, downscaled from $2048\times1360$. The edit is sky and exposure; structure is preserved.}
\label{fig:subsetqual}
\end{figure}

\paragraph{Sanity checks on a 20-image subset.}
Before the full run we trained on 16 pairs and held out 4, evaluating the base runs every 250 steps at $2048\times1360$ with 28 sampling steps (\cref{tab:overfit}, \cref{fig:subset,fig:subsetqual}). Three things are visible at this scale. First, the widened layer is used. We track $\rho$, the mean absolute weight of the new source half of the input layer divided by that of the pretrained half. It starts at 0, passes 1 between steps 1{,}250 and 2{,}500 in every base run, and passes 2 by step 5{,}000 in the longest run (B). Second, training half the time on native-resolution crops, together with conditioning dropout, gives the best validation PSNR and LPIPS of any subset run: 0.55 dB and 0.015 LPIPS better than run A, which trained on whole frames only and also used a constant learning rate instead of warmup and cosine decay, so the two differ in schedule as well as in crops. Third, in every base run validation PSNR peaks within the first 2K steps from the pretrained weights and is flat or falling afterwards, while the training loss keeps dropping (\cref{fig:subset}d). Under crops the training loss stays about five times higher than on whole frames by step 5{,}000 (0.60 against 0.12, 50-step mean), because half the steps see random windows the model cannot memorise. (Run B continues from A's checkpoint and never beats A's peak.) On 16 pairs the model memorises rather than generalises, so we do not read these numbers as a measure of editing quality. The distilled checkpoint at 512 pixels reaches a lower training loss than the base runs (median of the last 200 micro-batches 0.009, against 0.014 and 0.034) with no better validation PSNR. 

\begin{table*}[t]
\centering\small
\setlength{\tabcolsep}{3pt}
\caption{Subset runs on FLUX.2 Klein 4B: 16 training pairs, 4 validation pairs; base runs evaluated every 250 steps at $2048\times1360$, the distilled run every 100 steps at 512. Steps is the last evaluated step. Best validation PSNR (dB) and LPIPS over training with the step at which each occurred; $\rho$ is the final ratio of mean absolute weight in the conditioning half of the input projection to the pretrained half. Distilled rows use a different resolution and are indicative only.}
\label{tab:overfit}
\begin{tabular}{llrrr@{\,}lr@{\,}lc}
\toprule
Checkpoint & Recipe & Res. & Steps & \multicolumn{2}{c}{PSNR$\uparrow$ @step} & \multicolumn{2}{c}{LPIPS$\downarrow$ @step} & $\rho$ \\
\midrule
distilled & LoRA, no concat, lr $10^{-4}$ & 512 & 3{,}900 & 17.23 & {\scriptsize 3100} & 0.366 & {\scriptsize 1300} & -- \\ 
base & \chcat{}, full FT (A) & 2048 & 4{,}750 & 17.32 & {\scriptsize 1000} & 0.332 & {\scriptsize 1250} & 1.97 \\ 
base & \chcat{}, full FT (B, resumed from A) & 2048 & 9{,}250 & 17.22 & {\scriptsize 9000} & 0.339 & {\scriptsize 6750} & 2.36 \\ 
base & \chcat{}, full FT, crops 0.5, dropout 0.05 & 2048 & 5{,}500 & \textbf{17.87} & {\scriptsize 2000} & \textbf{0.317} & {\scriptsize 3500} & 1.23 \\ 
\bottomrule
\end{tabular}
\end{table*}

\subsection{What the widened input layer learned}
\label{sec:inputlayer}
The recipe's only architectural change is the widened input projection $W'=[\,W_{\mathrm{pre}}\;W_{\mathrm{cond}}\,]$, so we ask what its new half learned and what the change cost the pretrained half. \Cref{tab:inputlayer} inspects three editors: the main \chcat{} run (\cref{sec:method}), a follow-up \chcat{} run from the same base that restricts the higher learning rate ($10^{-4}$) to the new half, leaves the pretrained half at the body rate, and weights the loss on changed regions three times (its role in the copying test is described below), and, as a cross-architecture control, an SD3-Medium (2B) editor trained with the same surgery at the same resolution; this is a separate checkpoint from the SD3.5-Medium baseline of \cref{sec:experiments}. Three quantities are reported: the Frobenius norm of $W_{\mathrm{cond}}$ relative to $W_{\mathrm{pre}}$, the fraction of $W_{\mathrm{cond}}$ that lies outside the column space of $W_{\mathrm{pre}}$, and the relative drift of the pretrained weights from the base checkpoint. The effective rank of $W_{\mathrm{cond}}$ is 117 of 128 in the main run and 119 in the follow-up run.

\begin{table}[t]
\centering\footnotesize
\setlength{\tabcolsep}{3pt}
\caption{The input projection after training. Ratio: $\|W_{\mathrm{cond}}\|_F/\|W_{\mathrm{pre}}\|_F$. Outside: fraction of $W_{\mathrm{cond}}$ outside the column space of $W_{\mathrm{pre}}$. Drift: relative change of the pretrained input half and of the transformer body from the base checkpoint. For the main run the abs-mean ratio $\rho$ of \cref{tab:ckpt} is 1.31 at this checkpoint; the Frobenius ratio here is 1.10.}
\label{tab:inputlayer}
\begin{tabular}{lcccc}
\toprule
Editor & Ratio & Outside & Drift $W_{\mathrm{pre}}$ & Drift body \\
\midrule
Klein, main run, step 17K      & 1.10 & 97\% & 87\% & 2--4\% \\ 
Klein, follow-up, step 1.5K    & 0.53 & 99\% & 13\% & 7--9\% \\ 
SD3-M control, step 10.2K & 1.16 & 99\% & 27\% & 25--45\% \\ 
\bottomrule
\end{tabular}
\end{table}

In the main run the new half is as large as the pretrained half, spans nearly the full 128 dimensions (rank 117), and lies almost entirely outside the pretrained column space: the network built a separate reading path for the source rather than a scaled copy of the noise path. Every one of the 32 VAE channels is read with similar strength: the column norms of the new half span 3.2 to 5.1 against 3.2 to 4.3 for the pretrained half, and none is near zero. 

\paragraph{Copying versus editing.}
Having a separate reading path does not by itself mean the source is used to \emph{edit}. We test this with counterfactual conditioning on held-out pairs: the flow-matching loss at fixed noise levels with the true source, with zeros, and with the target itself in the conditioning channels. Averaged over three pairs, loss with the source is lower than with zeros only at the highest noise level ($\sigma=0.95$: 0.54 against 0.83); at $\sigma\le0.7$ the source is worse than zeros (at $\sigma=0.5$: 0.96 against 0.77), while the target in the conditioning channels lowers the loss two and a half to five times relative to zeros over $\sigma=0.3$ to $0.7$ (0.19 at $\sigma=0.5$). 

The follow-up run of \cref{tab:inputlayer}, which weights the changed-region loss three times and uses the separate input-layer learning rate, does not fix it. At steps 3{,}000 and 5{,}000 of that run the changed-region loss with the source remains above the loss with zeros (0.978 against 0.910, and 0.976 against 0.915; the main run at step 19{,}500 gives 0.988 against 0.901), the pull toward the source is unchanged (cosine 0.26, 0.24, 0.27), and the prompt still moves the output by under 4\%. Reweighting the loss makes the model copy slightly less blindly (loss with the target in the conditioning channels 0.41--0.43 against 0.33) without converting that into the edit. We report this as a negative result. Its checkpoints were also scored on the held-out set, but with text guidance 4 and image guidance 1.5, unlike every other Flux number in this paper, so we do not compare them to \cref{fig:heldout}. A second follow-up with a higher learning rate, flips and photometric augmentation was still training at submission and is not reported. 

\paragraph{Cross-architecture control.}
The SD3-Medium control behaves differently on the same tests. There the source helps: loss with the source is 2.5 to 3.9 times lower than with zeros at $\sigma\ge0.7$ and 1.6 times lower at $\sigma=0.5$, ties at $\sigma=0.3$, and hurts only at $\sigma=0.1$; the cosine with the copy direction is negative at high noise ($-0.23$ at $\sigma=0.95$) and turns positive only at $\sigma\le0.5$. The two architectures also fail in opposite directions: SD3-Medium overshoots the editor's sky and lands closer to the source by LPIPS despite doing more work, while Klein under-drives the edit. Both ignore prompt content outside the training tag vocabulary, which we attribute to the captions being permutations of four tags. One difference is procedural: the SD3 trainer parsed the conditioning-dropout flag but never applied it, and that editor produces a flat grey image when the source is zeroed, whereas the Klein editor, trained with dropout 0.05, draws a plausible scene; this is why the recipe keeps conditioning dropout. 

\section{Limitations}
\label{sec:limitations}

\textbf{Private data and weights.} The training set is private professional photography. We cannot release images or weights. We release the code, the full recipe, hyperparameters, and the evaluation protocol in the \textsc{Latent Fusion} repository.

\textbf{Single domain, single edit type.} All training and evaluation images are wide-field and close-range scenes from one editing workflow, and every edit is structure-preserving. We make no claim about content-changing edits, where a free source-target correspondence may be exactly what is needed, and the Kontext ranking may well hold.

\textbf{Scale.} About 1{,}400 pairs is small. Our findings may not hold for a model trained on a large edit dataset, where sequence concatenation has enough data to learn the correspondence.

\textbf{Evaluation.} Our metrics are full-reference PSNR, SSIM, and LPIPS against a single human-produced edit per image. There are many good edits of the same image, so a good output can still score poorly against the one reference we have. All outputs of the main run are sampled without guidance; the follow-up run was scored on the held-out set only with guidance (\cref{sec:ablations}), so it is not compared against it. We did not evaluate the follow-up unguided, so the region loss is not isolated from guidance. The two-pass guidance cost stated in \cref{sec:surgery} is the cost of the scheme, not a setting we swept.

\textbf{Compute.} Fine-tuning every weight of a 4B model at 10{,}880 tokens is expensive; the configuration is in \cref{sec:method}. The \chcat{} run cost roughly 680 GPU-hours. The \seqcat{} run has no surviving training log, so its cost and exact step count are not recorded, and the comparison in \cref{sec:results} rests on the evaluated checkpoints of a single run, which is not a verdict on the scheme. The pretrained \seqcat{} sweep in \cref{sec:results} is one frame without a reference edit. We did not sweep the design space; each ablation row is a single run.

\section{Conclusion}
\label{sec:conclusion}

We explored channel-wise concatenation on FLUX.2 Klein 4B base as the lower-cost way of feeding a source image to a transformer editor, in the regime where the edit must leave structure and identity alone and fewer than 1{,}500 pairs are available. Stacking the source onto the channels fixes the source-target correspondence by construction, keeps the sequence at its original length, and needs only two-pass classifier-free guidance where the InstructPix2Pix scheme needs three; appending it as tokens leaves that correspondence to be learned by attention over a sequence twice as long. In our runs, one of each scheme, the first kept the scene in place and the second moved it; we report that as an observation on this data, not as a verdict on sequence concatenation, which remains the design of choice for content-changing edits. The recipe is one widened input layer with a zero-initialised half and its own learning rate, training on native-resolution crops mixed with downscaled frames under conditioning dropout, and about 1{,}400 images. Whether sequence concatenation catches up with more data, as the Kontext report suggests it should, is a question we leave open. We hope the failure record in \cref{sec:appendix} saves others time.

{\small
\bibliographystyle{ieeetr}
\bibliography{refs}
}

\clearpage
\appendix
\onecolumn
\begin{multicols}{2}
\section{Lessons from training}
\label{sec:appendix}

Each item below cost us days. All are from our own runs; none are measured claims about other setups.

\paragraph{Vertical stripes are learned, not decoded.}
Our SD3 2K editor produced regular vertical stripes. We encoded and decoded the ground-truth images through the same VAE and got clean images back, so the VAE was not the cause. The transformer had learned to produce the stripes.

\paragraph{\texttt{-{}-scale\_lr} multiplies the learning rate.}
This Diffusers-style flag multiplies the learning rate by the number of GPUs, the per-GPU batch size, and the number of accumulation steps. Two SD3.5-M runs configured at $2\times10^{-6}$ trained at $2\times10^{-5}$ and $4\times10^{-5}$ (factors of 10 and 20 from 2 GPUs, batch 1 or 2, accumulation 5) and both crashed. 
Check the effective learning rate logged to wandb, not the flag.

\paragraph{Precision.}
All reported runs compute in bf16; the Flux runs keep a full-precision fp32 copy of the weights for the optimiser.

\paragraph{Resize to clean ratios.}
Shrinking sources by a factor that was not a whole number produced resampling artifacts, and the model learned to reproduce them. We now keep every training and inference resolution at an exact multiple of the 16-pixel token size, and place crop windows on that grid too.

\paragraph{The sky shortcut.}
For sky replacement the model learned ``sky is at the top of the frame'' rather than reading the image. Drone shots and low-horizon frames broke. Adding negative examples (frames where the top of the image is not sky and must not change) removed the shortcut.

\paragraph{The stock pipeline cannot sample a channel-wise concatenation checkpoint.}
The Flux inference pipeline appends the conditioning image as tokens, which is the sequence-concatenation layout. It cannot drive a widened input layer, so the trainer skips its built-in final inference for channel-wise concatenation runs. We sample those checkpoints with our own denoising loop; the stock pipeline is used only for the \seqcat{} baseline. 

\paragraph{Loss logging.}
Our SD3-era trainer logs two losses: \texttt{train\_loss}, averaged over GPUs and accumulation steps, and \texttt{step\_loss}, the raw value for one micro-batch. Our Flux trainer logs only the raw micro-batch loss, indexed by optimiser step, so with $k$ accumulation steps there are $k$ points at each x-value. Curves from different trainers are not comparable, and we only compare like with like. We also log $\rho$, the ratio of mean absolute weight in the new half of the input layer to the pretrained half; it starts at zero and shows whether the source channels are being used.

\paragraph{Rotary position embeddings.}
RoPE, how it behaves when the frame grows, and how it relates to \chcat{} cost us the most time to understand; \cref{sec:rope} collects it.

\section{Rotary position embeddings: what they do and how they scale}
\label{sec:rope}

\paragraph{The formula.}
A transformer sees its input as an unordered set of tokens; position has to be added. Rotary position embeddings (RoPE)~\cite{su2021roformer} add it by rotating the query and key vectors. Split a $d$-dimensional query $q$ into $d/2$ pairs of channels. For a token at position $m$, pair $j$ is rotated by the angle $m\theta_j$,
\begin{equation}
R(m\theta_j) =
\begin{pmatrix}
\cos m\theta_j & -\sin m\theta_j \\
\sin m\theta_j & \cos m\theta_j
\end{pmatrix},
\qquad
\theta_j = b^{-2j/d},
\end{equation}
for $j=0,\dots,d/2-1$ and a fixed base $b$. Writing $R_m$ for the block-diagonal matrix of all $d/2$ rotations, the attention score between a query at position $m$ and a key at position $n$ is
\begin{equation}
\label{eq:rope}
(R_m q)^\top (R_n k) = q^\top R_{n-m}\, k ,
\end{equation}
because two rotations compose into one by the difference of their angles. The score therefore depends only on the \emph{relative} position $n-m$.

Images need more than one coordinate. FLUX.2 splits the 128-dimensional head into four chunks of $d=32$ and rotates each by one coordinate of the token's position $p=(t,h,w,l)$: an image index $t$, the row $h$, the column $w$, and a layer index $l$, all with base $b=2000$. \Cref{eq:rope} holds per chunk, so a score depends on the offset $\Delta p = (\Delta t, \Delta h, \Delta w, \Delta l)$. The generated image's tokens get $p=(0,h,w,0)$.

\paragraph{How often each pair repeats.}
Pair $j$ comes back to its starting angle when $m\theta_j = 2\pi$, that is after
\begin{equation}
\lambda_j = 2\pi\, b^{2j/d}
\end{equation}
tokens. One token covers $16\times16$ pixels, so $16\lambda_j$ is the same distance in pixels. \Cref{tab:rope_wavelengths} gives $\lambda_j$ along one spatial axis of Klein. A $2048\times1360$ frame is 128 tokens wide and a $4096\times2720$ frame is 256.

\begin{center}
\small
\begin{tabular}{rrrrr}
\toprule
& \multicolumn{2}{c}{$b=2000$} & \multicolumn{2}{c}{$b=4189$ (NTK)} \\
\cmidrule(lr){2-3}\cmidrule(lr){4-5}
$j$ & tokens & pixels & tokens & pixels \\
\midrule
0 & 6.3 & 101 & 6.3 & 101 \\
4 & 42 & 672 & 50 & 809 \\
8 & 281 & 4{,}496 & 407 & 6{,}507 \\
12 & 1{,}879 & 30{,}066 & 3{,}272 & 52{,}348 \\
15 & 7{,}814 & 125{,}030 & 15{,}629 & 250{,}061 \\
\bottomrule
\end{tabular}
\captionof{table}{Tokens per full turn, $\lambda_j$, for pairs of one spatial RoPE axis of FLUX.2 Klein ($d=32$), at the pretrained base and at the NTK base used for 4K inference.}
\label{tab:rope_wavelengths}
\end{center}

\paragraph{Fast and slow pairs.}
The low-index pairs turn quickly. Pair 0 completes a turn every 6.3 tokens, so it cannot tell a token 1 step away from one 7.3 steps away; it is useful only for telling apart immediate neighbours, which is what local detail needs. The high-index pairs turn slowly. From $j=8$ upward a pair does not complete a turn across a whole 2K frame, so its angle changes little between any two tokens in the image and it acts as a coarse, long-range signal; the query-key match on those channels is then decided by content rather than by position. In an editor this suggests fast pairs for detail and slow pairs for layout.

\paragraph{Train short, infer long.}
A model trained at one sequence length and run at a longer one sees offsets $\Delta p$ it never saw in training: at $4096\times2720$ a token's neighbours reach up to 255 positions away, against 127 at 2K. The common fixes slow the rotations down so the new offsets map back onto angles seen in training. Let $s$ be the length ratio, here $s=2$ per spatial axis. Position interpolation (PI) divides every angle by $s$, $\theta'_j = \theta_j/s$, which also blurs the fast pairs that local detail relies on. NTK-aware scaling instead raises the base,
\begin{equation}
b' = b\, s^{d/(d-2)}, \qquad \theta'_j = \theta_j\, s^{-2j/(d-2)},
\end{equation}
so the fastest pair is left alone and only the slowest is slowed by the full $s$; for Klein, $b'=2000\cdot2^{32/30}\approx4189$. YaRN splits the pairs into three bands: fast pairs are untouched, slow pairs are divided by $s$, and the pairs between them blend the two, $\theta'_j = (1-\gamma_j)\,\theta_j/s + \gamma_j\,\theta_j$ with $\gamma_j$ falling from 1 to 0 across the band. \Cref{fig:rope_scaling} compares the three.

\begin{center}
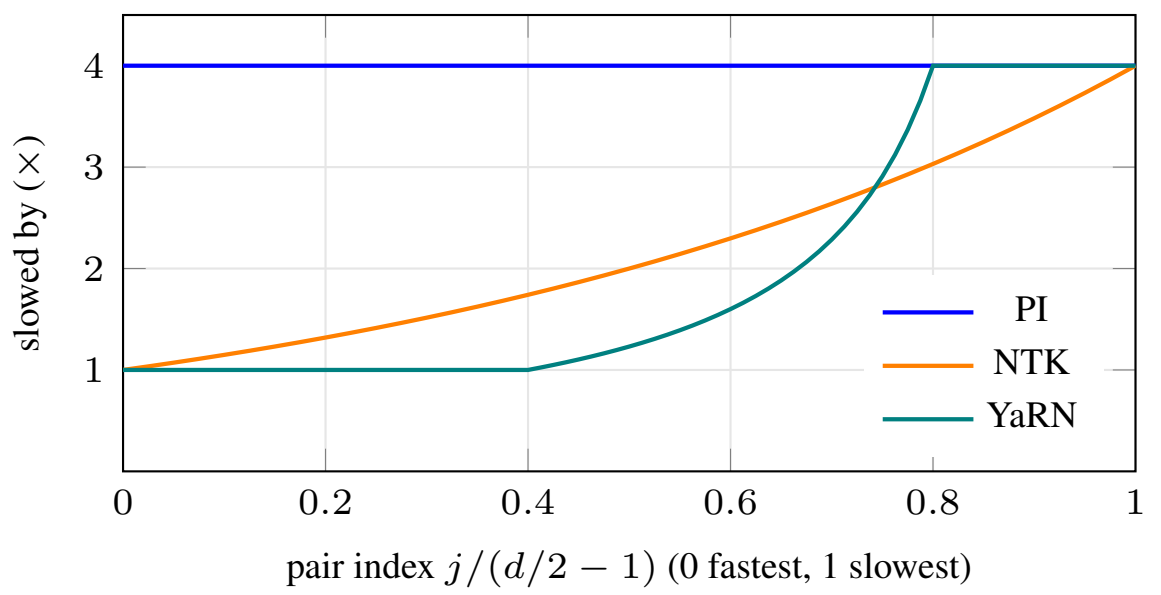

\begin{tikzpicture}
\begin{axis}[
  width=0.95\columnwidth, height=4.6cm,
  xlabel={pair index $j/(d/2-1)$ (0 fastest, 1 slowest)},
  ylabel={slowed by ($\times$)},
  xmin=0, xmax=1, ymin=0, ymax=4.5,
  ytick={1,2,3,4},
  legend pos=south east, legend style={font=\scriptsize, draw=none, fill=white},
  tick label style={font=\scriptsize}, label style={font=\scriptsize},
  grid=major, grid style={gray!20},
]
\addplot[blue, thick, domain=0:1, samples=2] {4};
\addlegendentry{PI}
\addplot[orange, thick, domain=0:1, samples=41] {4^x};
\addlegendentry{NTK}
\addplot[teal, thick, domain=0:1, samples=81]
  {1/((1-max(0,min(1,(0.8-x)/0.4)))/4 + max(0,min(1,(0.8-x)/0.4)))};
\addlegendentry{YaRN}
\end{axis}
\end{tikzpicture}
\captionof{figure}{How much each RoPE pair is slowed under PI, NTK and YaRN, for an illustrative length ratio $s=4$; the YaRN ramp from 0.4 to 0.8 is also illustrative. PI slows every pair equally; NTK slows pairs progressively; YaRN keeps the fast pairs exact, slows the slow pairs fully and blends the middle band.}
\label{fig:rope_scaling}
\end{center}

\paragraph{What we found at 4K.}
We ran the main $2048\times1360$ \chcat{} checkpoint at $4096\times2720$ inference. With NTK scaling the base went from 2000 to 4189. Turning RoPE off ruins the image, so the network depends on it. NTK scaling and attention sharpening, which scales the attention logits up so that four times as many keys do not flatten the softmax, barely changed the output: at this ratio RoPE copes with 4K frames as it is. Our tests agree with the picture above, with the fast pairs carrying detail and the slow pairs carrying structure. At 4K each query attends over 43{,}520 tokens instead of 10{,}880, and this attention dilution mostly weakens the edit rather than the detail copied from the source. That is what \chcat{} predicts: the source detail reaches each token through its own input channels, not through attention, so spreading attention thinner leaves it in place.

\paragraph{How \seqcat{} uses RoPE.}
Under \seqcat{} the source tokens are appended to the sequence and need positions of their own. The Klein pipeline gives the $i$-th reference image $p=(10+10i,\,h,\,w,\,0)$: the same rows and columns as the target, shifted on the image axis. For a target token and the source token at the same pixel block, the offset is $\Delta p = (10,0,0,0)$. Nothing else marks that token as the one to copy from. The network has to learn, through attention over all $2L$ tokens, that the useful source token is the one with $\Delta h=\Delta w=0$, and to keep looking there at every layer. With enough data this is learned; with about 1{,}400 pairs our \seqcat{} run let the scene drift (\cref{sec:results}), which is what a correspondence that is only softly learned would look like.

\paragraph{Carrying InstructPix2Pix over to a transformer.}
InstructPix2Pix~\cite{brooks2023instructpix2pix} stacks the source latent onto the noisy latent along the channel axis of a U-Net. There, alignment comes for free: the first convolution reads both stacks at the same pixel. The question for a transformer was what plays the role of that shared pixel. The answer is the token position. If the source block is stacked into the same token as the target block, as $[\,x_{t,i} \,\|\, c_i\,]$, both inherit the one position $p_i=(0,h_i,w_i,0)$, and their offset is $\Delta p = 0$ by construction. No new positions are created, no RoPE axis is offset, the sequence stays at $L$, and the pretrained position grid is reused unchanged. The only new parameters are the zero-initialised half of the input projection (\cref{sec:surgery}).

The lesson we draw is narrow. RoPE only tells a token how far away another token is; it does not tell it which one to copy from. When source and target are pixel-aligned, that choice is known in advance, and putting both in the same token states it directly instead of leaving attention to discover it. When the edit moves content, the choice is not known in advance, and the freedom of \seqcat{} is what the task needs.

\end{multicols}

\end{document}